\documentclass[conference]{IEEEtran}
\usepackage{geometry}
\usepackage{xurl} 
\usepackage[bookmarks=false]{hyperref}
\usepackage{amsmath,amsfonts}
\usepackage{array}
\usepackage[noadjust]{cite}
\usepackage{textcomp}
\usepackage{stfloats}
\usepackage{url}
\usepackage{verbatim}
\usepackage{graphicx}
\usepackage{siunitx}
\DeclareSIUnit\mt{\milli\tesla} 
\usepackage[utf8]{inputenc}
\usepackage[english]{babel}
\usepackage[linesnumbered,ruled,lined]{algorithm2e}
\usepackage[noadjust]{cite}
\usepackage{multirow} 
\usepackage{setspace}
\usepackage{subcaption}
\usepackage{algcompatible}
\usepackage{array}
\usepackage{amsthm}
\usepackage{adjustbox}
\usepackage{bbm}
\let\oldnl\nl
\newcommand{\nonl}{\renewcommand{\nl}{\let\nl\oldnl}}
\newcolumntype{L}[1]{>{\raggedright\let\newline\\\arraybackslash\hspace{0pt}}m{#1}}
\newcolumntype{C}[1]{>{\centering\let\newline\\\arraybackslash\hspace{0pt}}m{#1}}
\newcolumntype{R}[1]{>{\raggedleft\let\newline\\\arraybackslash\hspace{0pt}}m{#1}}

\algnewcommand\KwEvl{\textbf{Evaluation:}}
\usepackage{booktabs}

\usepackage{makecell}
\usepackage{amsmath}
\usepackage{siunitx}
\usepackage{caption} 

\usepackage[T1]{fontenc}
\usepackage{textcomp}
\usepackage{newtxtext,newtxmath}    
\usepackage[scaled=0.85]{beramono}  

\begin{document}

\title{Causal Beam Selection for Reliable Initial Access in AI-driven Beam Management}

\author{
    \IEEEauthorblockN{
        Nasir Khan\IEEEauthorrefmark{1}, 
        Asmaa Abdallah\IEEEauthorrefmark{2}, 
        Abdulkadir Celik\IEEEauthorrefmark{3}, 
        Ahmed M. Eltawil\IEEEauthorrefmark{2}, 
        Sinem Coleri\IEEEauthorrefmark{1}
    }
    \IEEEauthorblockA{\IEEEauthorrefmark{1}Department of Electrical and Electronics Engineering, Koç University, Istanbul, Turkey}
    \IEEEauthorblockA{\IEEEauthorrefmark{2}CEMSE Division, King Abdullah University of Science and Technology (KAUST), Thuwal, Saudi Arabia}
    \IEEEauthorblockA{\IEEEauthorrefmark{3}School of Electronics and Computer Science, University of Southampton, Southampton, U.K.}
    \IEEEauthorblockA{Emails: \{nkhan20, scoleri\}@ku.edu.tr, \{asmaa.abdallah, ahmed.eltawil\}@kaust.edu.sa, a.celik@soton.ac.uk}
}

\maketitle

\begin{abstract}  
Efficient and reliable beam alignment is a critical requirement for mmWave multiple-input multiple-output (MIMO) systems, especially in 6G and beyond, where communication must be fast, adaptive, and resilient to real-world uncertainties. Existing deep learning (DL)-based beam alignment methods often neglect the underlying causal relationships between inputs and outputs, leading to limited interpretability, poor generalization, and unnecessary beam sweeping overhead. In this work, we propose a causally-aware DL framework that integrates causal discovery into  beam management pipeline. Particularly, we propose a novel two-stage causal beam selection algorithm to identify a minimal set of relevant inputs for beam prediction. First, causal discovery learns a Bayesian graph capturing dependencies between received power inputs and the optimal beam. Then, this graph guides  causal feature selection for the DL-based classifier. Simulation results reveal that the proposed causal beam selection matches the performance of conventional methods while drastically reducing input selection time by 94.4\% and beam sweeping overhead by 59.4\% by focusing only on causally relevant features.

\end{abstract}

\IEEEpeerreviewmaketitle

\section{Introduction}
\noindent  \IEEEPARstart{T}{he}  vision of AI-native 6G extends 5G by embedding artificial intelligence (AI) across all communication layers to enable intelligent and adaptive system design. Leveraging millimeter-wave (mmWave) frequencies, 6G promises ultra-high data rates but faces significant challenges in beam management (BM) due to severe propagation loss. Directional transmission is necessary, making high beamforming gain and minimal training overhead critical. BM entails identifying and maintaining optimal beam pairs between the base station (BS) and user equipment (UE), where beam sweeping is central to initial access (IA) and requires frequent measurements to preserve link reliability under mobility and blockage~\cite{tutorialBA}. These complexities highlight the need for intelligent, adaptive BM strategies to ensure robust and fast mmWave communications.

Traditional exhaustive beam sweeping over predefined codebooks, such as discrete Fourier transform (DFT) codebooks, incurs high overhead, particularly with large antenna arrays. To mitigate this signaling overhead, AI-driven BM has emerged as a promising solution, enabling beam prediction from limited sensing data to reduce training overhead and enhance adaptability. Current 3GPP efforts aim to standardize AI/Machine learning (ML) for beam prediction to lower measurement complexity. Despite improved alignment efficiency, existing AI/ML methods lack transparency and systematic input feature selection, which are crucial for reliable and trustworthy beam management \cite{our_work}.


The accuracy of AI/ML-based beam classifiers depends not only on the number of input beams but also on their specific combinations, making input selection a key feature selection task. Traditional (non-causal) methods evaluate features using statistical metrics such as correlation, mutual information, or model-derived importance scores. Features are labeled as \emph{strongly relevant}, \emph{weakly relevant}, or \emph{irrelevant}, with selection aiming to retain the most relevant ones. These methods rank and iteratively select features for optimal performance.  Post-hoc explainable AI (XAI) methods for feature selection have been used to identify key attributes that influence model decision-making and explain predictions of trained models \cite{XAI1,  XAI4, XAI_ICC}. Among the post-hoc XAI methods,  feature attribution techniques, such as Shapley Additive Explanations (SHAP) \cite{lundberg}, can highlight the features that most impact the model's predictions. Along this line of research, classical XAI methods and SHAP have been applied to identify and interpret wireless KPIs in 5G network slicing \cite{XAI1}  network intrusion detection  \cite{XAI4} and radio resource management \cite{kHAN-TCOM}, \cite{XAI_ICC}. However, feature importance ranking  is coupled with the availability of large-scale wireless datasets and pre-trained models requiring enormous computation time. Existing  AI/ML models for beam management tasks are often computationally intensive, model-dependent and  concentrate on statistical correlations within the training data, overlooking the features that are causally relevant.  Integrating  causal reasoning can address these gaps, enhancing interpretability, generalization, and robustness by reducing the model’s exposure to misleading inputs.

Recent studies attribute inefficiencies in BM partly to the absence of causal formalism in current AI/ML-based system design and data generation processes~\cite{causal_magazine}. Causal feature selection, inspired by Bayesian networks, aims to identify features with direct or indirect causal influence on the target variable, helping models focus on inputs that truly affect outcomes. Causal discovery methods uncover such relationships from observational wireless data using directed acyclic graphs (DAGs), where nodes represent variables and edges denote causal influence  ~\cite{causal_discovery_survey}. These methods have been applied in joint communication and sensing~\cite{causal1}, semantic communication \cite{causal3}, and 6G automation~\cite{causal4} to enhance decision-making by identifying causal links among system variables and KPIs. However, prior works focus on semantic-level generalization, while our approach introduces a systematic causal input selection methodology tailored to BM. This enables reduced input dimensionality, lower beam sweeping overhead, and improved interpretability, with significantly less computation compared to traditional~\cite{FastIA } and XAI-based methods~\cite{KHAN_TWC}.

In this paper, a deep learning (DL)-based beam alignment engine (BAE) is proposed that utilizes the received signal strength
indicators (RSSIs) measurements from a finite set of sensing beams (DFT codebook) to predict optimal narrow beams from the oversampled DFT (O-DFT) codebook for IA and data transmission. We propose a novel two-stage framework grounded in causal analysis to identify a minimal subset of causally relevant inputs for beam prediction. In the first stage, causal discovery is used to learn the optimal Bayesian graph structure that captures dependencies among model inputs (received power measurements) and the optimal beam. In the second stage, the learned graph is leveraged for causal feature selection, identifying the input features most relevant to the DL-based BAE. The selected features closely align with those obtained via SHAP-based feature selection while significantly reducing the computational overhead typically required by SHAP.

\begin{figure}
    \centering
    \includegraphics[width=1\linewidth]{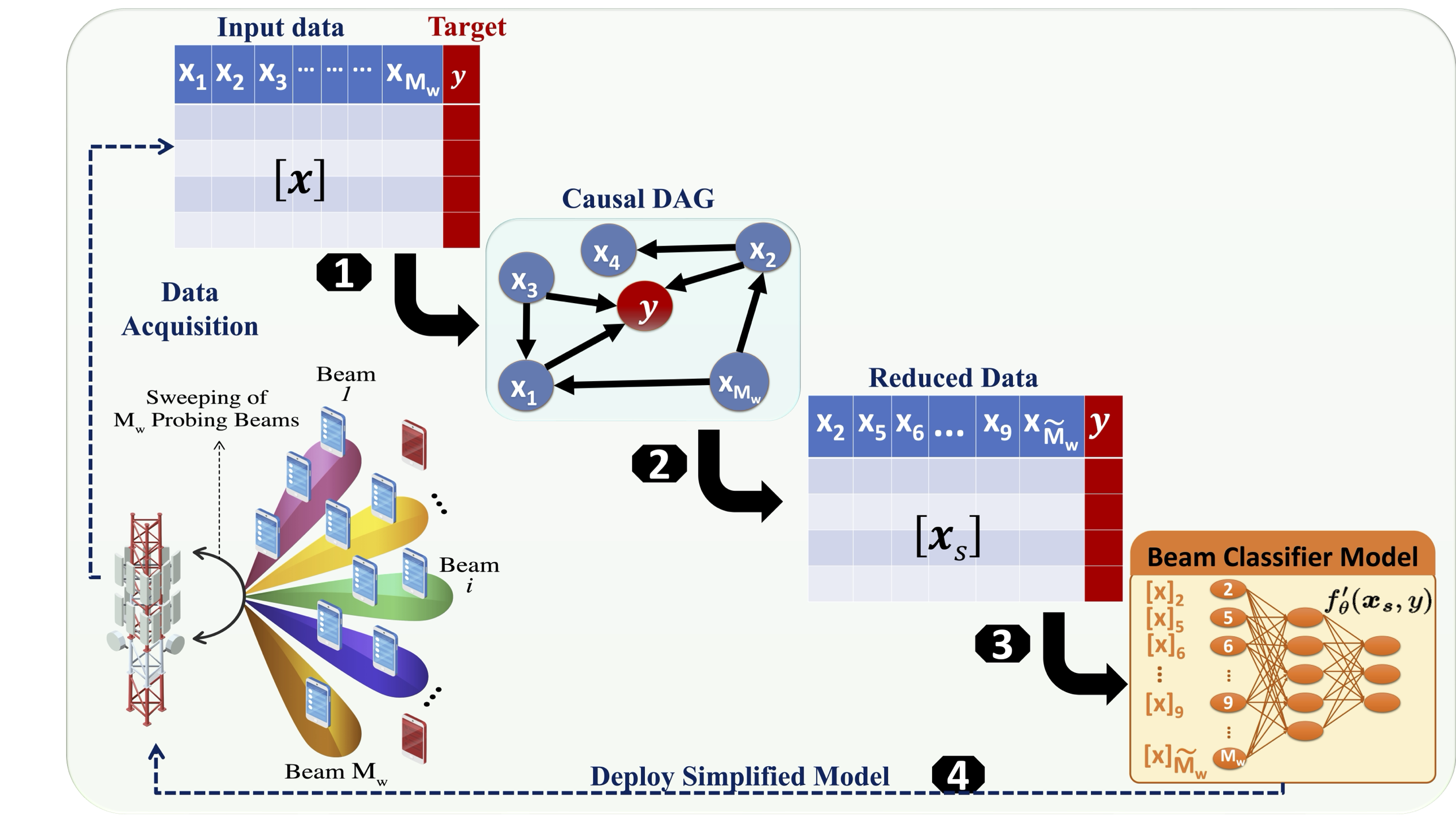}
    \setlength{\belowcaptionskip}{-10pt}
    \caption{Illustration of the system model and the proposed causal beam selection strategy.}
    \label{proposed_flow}
\end{figure}

\section{System Model} \label{sec:system}
We consider a narrowband downlink mmWave system where the BS, equipped with a uniform linear array (ULA) of $N_{\mathrm{BS}}$ antennas, serves $N_\mathrm{U}$ single-antenna UEs, as shown in Fig.~\ref{proposed_flow}. The scenario involves multi-user beamforming, where each UE receives a single stream. At mmWave frequencies, channels are typically sparse in the angular domain, resulting in a limited number of paths $L$. The channel from the BS to $\mathrm{UE}_u$ follows a geometric multipath model:
\begin{equation}
    \mathbf{h}_u =  \alpha_{u,l} \mathbf{b}\left(\phi_{u,l}\right),
\end{equation}
where $\alpha_{u,l}$ is the complex path gain, $\phi_{u,l}$ is the angle of departure (AoD), and $\mathbf{b}(\phi_{u,l})$ is the beam steering vector defined as
\begin{equation}
    \mathbf{b}(\phi_{u,l}) = \frac{1}{\sqrt{N_{\mathrm{BS}}}} \left[1, e^{j \frac{2 \pi}{\lambda} d \sin(\phi_{u,l})}, \dots, e^{j (N_{\mathrm{BS}} - 1) \frac{2 \pi}{\lambda} d \sin(\phi_{u,l})}\right]^T,
\end{equation}
where $\lambda$ is the wavelength and $d = \lambda/2$ is the antenna spacing.

To reduce hardware complexity, we employ analog-only beamforming with a single RF chain shared across $N_{\mathrm{BS}}$ antennas. The beamforming vector is given by
\begin{equation}
\mathbf{w} = \frac{1}{\sqrt{N_{\mathrm{BS}}}} \left[e^{j \varphi_1}, \ldots, e^{j \varphi_i}, \ldots, e^{j \varphi_{N_{\mathrm{BS}}}}\right]^{\top} \in \mathbb{C}^{N_{\mathrm{BS}} \times 1},
\end{equation}
where $\varphi_i$ is the phase of the $i$-th antenna element.

The BS uses a beamforming codebook $\mathbf{W} = \{\mathbf{w}_1, \ldots, \mathbf{w}_Y\}$ with $Y$ predefined vectors, each satisfying $\|\mathbf{w}_y\|^2 = 1, \forall y$, to meet the constant-modulus constraint.

During beam sweeping, the BS transmits symbols $s_w \in \mathbb{C}$ through beams in $\mathbf{W} \in \mathbb{C}^{N_{\mathrm{BS}} \times Y}$. The received signal at $\mathrm{UE}_u$ via the $q_u$-th beam $\mathbf{w}_{q_u}$ is
\begin{equation}
   r_u = \sqrt{P_{\mathrm{BS}}} \mathbf{h}_{u}^H \mathbf{w}_{y_u} s_w + z_u,
\end{equation}
where $P_{\mathrm{BS}}$ is the transmit power, $\mathbf{h}_u \in \mathbb{C}^{N_{\mathrm{BS}} \times 1}$ is the channel vector, and $z_u$ is additive noise with power $\sigma_z^2$. Assuming unit-power symbols, the SNR at $\mathrm{UE}_u$ is given by
\begin{equation}
    \text{SNR}_u = \frac{P_{\mathrm{BS}} \left| \mathbf{h}_{u}^H \mathbf{w}_{y_u} \right|^2}{\sigma_z^2}.
\end{equation}

\section{Problem Formulation and Solution Methodology}\label{problem_outline}
This section describes the problem formulation for the beam alignment task, the data acquisition approach and the DL-based beam alignment procedure. 
\subsection{Problem Formulation}

For a given BS-UE pair, the optimal beam index $ y_{u}^*$ can be identified by selecting the beam that maximizes the SNR:
\begin{align} \label{beam_selection}
    y_{u}^* &= \underset{y_u \in \left\{1,2,\dots,Q\right\}}{\arg \max} 
    \left( \frac{P_{\mathrm{BS}}\left| \mathbf{h}_{u}^H \mathbf{w}_{y_u} \right|^2 }{\sigma_z^2} \right) \notag \\
    &= \underset{y_{u{}} \in \left\{1,2,\dots,Q\right\}}{\arg \max} 
    \left( \left| \mathbf{h}_{u}^H \mathbf{w}_{y_u} \right|^2 \right).
\end{align}

Importantly, the total number of beams swept remains constant, regardless of the number of UEs, since multiple UEs can simultaneously measure sensing beams using shared time-frequency resources. However, identifying $y_{u}^*$ via exhaustive search over $Y$ codewords incurs high sweeping overhead. To mitigate this, we exploit deep learning's exploration ability to predict optimal beam directions for each site.

\subsection{Data Acquisition}\label{DA_subsection}
As shown in Fig. \ref{proposed_flow}, the proposed solution leverages the RSSI values over a small set of compact wide sensing beams $M_{\mathrm{w}}$ (i.e., beams from DFT codebook with $M_{\mathrm{w}}\ll Y$ ), to select the best narrow beam from the O-DFT codebook of size $Y$, thus, avoiding exhaustive extensive  search over the O-DFT codebook. For beam-sweeping, the BS transmits the pilot signals over a smaller set of beams, each at a separate time slot.  All UEs connected to the BS measure and report the RSSI values of the $M_{\mathrm{w}}$ beams. It is assumed that beam sweeping, measurement, and reporting occur within the coherence time during which the channel remains constant. The reported beam sweeping results for  UE  $u$ can be written as
\begin{equation}\label{rp}
\mathbf{x}_u={\left[[x_u]_1,\ldots,[x_u]_{M_\mathrm{w}} \right]}=\left[\left|\left[r_{u}\right]_1\right|^2 \cdots\left|\left[r_{u}\right]_{M_{\mathrm{w}}}\right|^2\right]^T ,
\end{equation}
where $\left[r_{u}\right]_i=\sqrt{P_{\mathrm{BS}}} \mathbf{h}_u^H \mathbf{w}_is_w+z_i$ is the received signal using the $i$-th sensing beam, $\forall i \in M_{\mathrm{w}}$.

The ground-truth label $y_u$ for each data sample $\mathbf{x}_u$ is obtained  by searching the optimal beam direction in the O-DFT codebook by performing noise-free exhaustive search. The constructed  dataset is denoted by \( \mathcal{D} = \{ (\mathbf{x}_u, y_u) \}_{u=1}^{D} \), where \( \mathbf{x}_u \in \mathbb{R}^{M_{\mathrm{w}}} \) represents the beam sweeping measurements for user \( u \),  \( y_u \in \{1, 2, \ldots, Y\} \) denotes the ground-truth label corresponding to the optimal beam index with the highest RSSI as the target label, generated using \eqref{beam_selection}, and  \( D \) denotes the total number of samples in the dataset.

\subsection{DL-based Beam Alignment}\label{DL_subsection}
In the proposed 3GPP-compliant BM framework, a DL-based beam prediction model is trained to associate the best beam between the BS and UE using RSSI measurements over wide sensing beams as input and the index of the optimal narrow beam as output. Particularly, the reported beam sweeping results from (\ref{rp}) are fed as input to a deep neural network (DNN)-based beam classifier denoted by \( f\left(\cdot \hspace{0.05cm}; \boldsymbol{\theta} \right): \mathcal{X} \rightarrow \mathbb{R}^Y \), where \( \mathcal{X} \in \mathbb{R}^{M_\mathrm{w}} \) is the DNN input space corresponding to RSSI values across \( M_{\mathrm{w}} \) beams, and \( \boldsymbol{\theta} \) denotes the model parameters.  Furthermore, cross-entropy loss, which quantifies the dissimilarity between predicted probabilities and true labels, is applied to train the DNN's weight parameters, using the one-hot encoding of the optimal beam index 
in \eqref{beam_selection} as the ground truth label.
The  loss function is formulated as: 
\begin{equation} \label{ref_loss}
\mathcal{L}(\boldsymbol{\theta} , \mathcal{D}) = - \frac{1}{D} \sum_{d=1}^{D} \sum_{P_u=1}^{Y} p_{u,d,y_u}^\star \log \widehat{p}_{u,d,y_u},
\end{equation} 
{where $p_{u,d, y_u}^\star$ is the true distribution over  labels indicating that the $d$-th data sample belongs to $y_u$-th class, such that $p_{u,d, y_u}^\star = 1$ if the $y_u$-th candidate narrow beam is the actual
optimal beam ($y_u^\star$) and $p_{u,d, y_u}^\star = 0$ otherwise, and $\widehat{p}_{u,d, y_u} \in$ $\mathbb{R} \rightarrow (0,1)$ represents the predicted probability distribution.


\section{CAUSALITY BASED BEAM SELECTION}\label{CFS_subsection}
 In this section, we treat beam selection as a causal feature selection problem, aiming to identify inputs that have a true causal influence on the target rather than relying on spurious correlations. This is achieved by modeling the data as a DAG, where each node represents either a input or the target beam index, and edges reflect potential causal dependencies. The goal is to isolate the subset of RSSI features, i.e, sensing beam measurements that have a direct causal influence on selected narrow beam, effectively filtering out redundant or spurious features.
 
 
\subsection{Causal Beam Discovery} \label{CBS}
The DL-based beam alignment strategy described in the previous section relies on sweeping sensing beams at the BS to collect RSSI values \eqref{rp}, which are used as input features to the DL model.  To reduces signaling overhead by avoiding sweeping the entire set of sensing beams and improve alignment efficiency, we model the  causal relationships between the input features and the  target variable using a DAG \( G = \{V, E\} \), where each node in \( V \) corresponds to a feature in the dataset, including the RSSI values \( \mathbf{x}_u \in \mathbb{R}^{M_{\mathrm{w}}} \) and the target beam index \( y_u \in \{1, \ldots, Y\} \). The directed edges \( E \) represent causal influences among these features. The goal of causal discovery is to estimate the causal structure  from observed samples \( \{ (\mathbf{x}_u, y_u) \}_{u=1}^{N} \). We assume that the target \( y_u \) is not a cause of any RSSI feature, i.e., \( (y_u, [x_u]_i) \notin E \), aligning with the unidirectional nature of the underlying physical process. To estimate the causal  DAG structure, we use the Direct Linear Non-Gaussian Acyclic Model (DirectLiNGAM) algorithm \cite{directlingam}, which assumes a linear non-Gaussian data-generating process. The method iteratively applies least squares regressions to compute residuals and selects the variable most independent from its residuals to build a causal ordering. This ordering is then used to estimate connection strengths between features. The resulting DAG captures directional dependencies among the dataset's features and the targe variable.

As shown in Fig.~\ref{proposed_flow}, the output of DirectLiNGAM is a causal graph where each node represents a dataset feature, and a directed edge \( x_i \xrightarrow{s} x_j \) indicates that feature \( x_i \) causes feature \( x_j \), with \( s \) denoting the estimated causal strength. The DirectLiNGAM algorithm estimates the underlying causal structure among input features and the target variable, assuming a linear non-Gaussian data-generating process. Given the dataset \( \mathcal{D} = \{ (\mathbf{x}_u, y_u) \}_{u=1}^N \), we define the joint variable 
\[
\mathbf{z} = [[x_u]_1, \dots, [x_u]_{M_{\mathrm{w}}}, y]^\top \in \mathbb{R}^p,
\]
where \( p = M_{\mathrm{w}} + 1 \). The causal generative model is formulated as:
\begin{equation}
\mathbf{z} = \mathbf{E} \mathbf{z} + \mathbf{e},
\end{equation}
where \( \mathbf{E} \in \mathbb{R}^{p \times p} \) is a strictly lower-triangular effect matrix encoding causal influences, and \( \mathbf{e} \in \mathbb{R}^p \) is a vector of mutually independent non-Gaussian disturbances. A nonzero element \( E_{ij} \neq 0 \) implies a directed edge \( z_j \rightarrow z_i \), corresponding to a causal relationship in the DAG.

The proposed Causal Beam Selection Algorithm is  summarized in Algorithm~\ref{alg:CFS}. The algorithm first estimates the effect matrix \( \mathbf{E} \) (Lines 1--2) using the DirectLiNGAM procedure, which iteratively regresses each variable in \(\mathbf{z}\) onto others and ranks them based on residual independence using a kernel-based metric, resulting in a causal ordering from which direct causal effects are estimated via least squares. Next, the direct parents of the target variable \( y \) are selected as all features \( i \) such that \( E_{y,i} \neq 0 \) (Lines 3--4). To ensure inclusion of other potentially relevant variables, the algorithm computes the total connectivity for each feature \( i \) (Lines 5--9), defined as the sum of its incoming connections (number of nonzero effects from other features to \( i \)) and outgoing connections (number of nonzero effects from \( i \) to other features), and ranks features accordingly (Line 11). Finally, the selected feature set \( \mathcal{S} \) is constructed by combining the direct parents of the target with top-ranked features based on total connectivity until \( |\mathcal{S}| =  \widetilde{M}_{\mathrm{w}} \)  (Lines 13--18). 

Algorithm~\ref{alg:CFS} yields a compact, causally motivated subset of features for robust and interpretable beam prediction. The inputs of the algorithm are the dataset \( \mathcal{D} =  \{ (\mathbf{x}_u, y_u) \}_{u=1}^{N} \) and maximum feature count \( F_{\max} \). The output is a subset of causally relevant RSSI measurements \( \mathbf{x}_S = \{x_i \mid i \in \mathcal{S} \} \), where \( \widetilde{M}_{\mathrm{w}} = |\mathcal{S}| \ll M_{\mathrm{w}} \). This subset enables a simplified model with reduced sensing and inference complexity. As shown in Fig \ref{proposed_flow}, a simplified model is then  trained using this subset of critical features $\mathbf{x}_S$, reducing model complexity (e.g., via feature selection) and beam sweeping time by minimizing the number of sensing beams to \( \widetilde{M}_{\mathrm{w}} \) instead of \( {M}_{\mathrm{w}} \). Note that unlike post-hoc XAI based feature selection strategies, the proposed DirectLiNGAM based causal beam selection approach does not require a pretrained model to infer the importance of selected features and is scalable to high-dimensional inputs \cite{directlingam}.


\begin{algorithm}[t] \footnotesize
\caption{Causal Beam Selection Algorithm}
\label{alg:CFS}

\SetKwInOut{Input}{Input}
\SetKwInOut{Output}{Output}

\Input{Dataset \( \mathcal{D} \), maximum feature count  \( \widetilde{M}_{\mathrm{w}} \)}
\Output{Subset of input features \( \mathbf{x}_S \)}

Train DirectLiNGAM on \( \mathcal{D} \) to obtain effect matrix \( \mathbf{E} \)\;

Derive causal ordering from residual independence\;

Select direct parents of \( y \):\;
\quad \( \mathcal{P}(y) \gets \{ i \mid E_{y,i} \neq 0 \} \)\;

\textbf{Compute total connectivity for each feature \( i \):}\;
\For{\( i \gets 1 \) \KwTo \( p \)}{
    \( \text{In}(i) \gets \sum_j \mathbf{1}_{E_{i,j} \neq 0} \)\;
    \( \text{Out}(i) \gets \sum_j \mathbf{1}_{E_{j,i} \neq 0} \)\;
    \( \text{Conn}(i) \gets \text{In}(i) + \text{Out}(i) \)\;
}

Sort features by \( \text{Conn}(i) \) in descending order\;

Initialize \( \mathcal{S} \gets \mathcal{P}(y) \)\;

\While{\( |\mathcal{S}| < F_{\max} \)}{
    Add next top-ranked feature by total connectivity to \( \mathcal{S} \)\;
}

Trim \( \mathcal{S} \) to size \( F_{\max} \): \( \mathcal{S} \gets \mathcal{S}[:  \widetilde{M}_{\mathrm{w}}  ] \)\;

Form reduced input: \( \mathbf{x}_S \gets \{x_i \mid i \in \mathcal{S} \} \)\;

\Return \( \mathbf{x}_S \)\;

\end{algorithm}

\section{Performance Evaluation}\label{sec:simulation}
In this section, we provide details of the simulation setup, and performance evaluation metrics, followed by a discussion of the results.
\subsection{Simulation Setup}

 We simulate mmWave BS-UE communication in an urban setting modeled after downtown Boston, using the  Boston5G scenario from DeepMIMO~\cite{alkhateeb2019deepmimo}, The BS  is equipped with a 32-element ULA at 15\,m height facing the negative $y$-axis, and single-antenna UEs at 2\,m. The $200\,\mathrm{m} \times 230\,\mathrm{m}$ area is discretized with $0.37\,\mathrm{m}$ UE spacing. Channel vectors $\mathbf{h} \in \mathbb{C}^{32 \times 1}$ are normalized by their maximum absolute value to improve training stability. The BS performs analog beamforming with $M_\mathrm{w}= 32$ sensing beams using a $N_{\mathrm{BS}}$-DFT codebook, whereas, for the narrow beam codebook $\mathbf{W}$, we use an O-DFT  with oversampling  factor of 4 to get a total of 128 narrow beams.
 In all experiments, 70\% of the data is allocated for training, 10\% for validation, and the remaining 20\% for testing. for the model architecture, we utilize a fully connected DNN with $L=3$ hidden layers and ReLU activations is used. The input layer matches the number of RSSI values, and the output layer has neurons equal to the possible O-DFT indices, producing a softmax distribution. Additional configurations and hyperparameters are listed in Table~\ref{tab:combined-hyperparams}.

\begin{table}[!t]
\centering
\caption{\textsc{Simulation and Training Parameters}} 
\label{tab:combined-hyperparams}
\begin{adjustbox}{width=0.65\columnwidth,center}
\footnotesize
\begin{tabular}{|l|l|} 
\hline
\multicolumn{2}{|c|}{\textbf{Channel Generation Parameters}} \\ \hline
Active BS                   & 1                                 \\ \hline
BS transmit power           & 30 dBm                            \\ \hline
Active users                & 1--622                            \\ \hline
Antenna configuration       & $(1,32,1)$ (ULA)                  \\ \hline
Carrier frequency           & 28 GHz                            \\ \hline
System bandwidth            & 500 MHz                           \\ \hline
Antenna spacing             & $0.5\lambda$                      \\ \hline
OFDM sub-carriers           & 1                                 \\ \hline
OFDM sampling factor        & 1                                 \\ \hline
OFDM limit                  & 1                                 \\ \hline
$D_{\text{train}}$ size     & 57144                             \\ \hline
\multicolumn{2}{|c|}{\textbf{DNN Architecture and Training}}  \\ \hline
Input layer                 & Neurons = \# RSSI values          \\ \hline
Hidden layer 1              & 64 neurons, ReLU                  \\ \hline
Hidden layer 2              & 64 neurons, ReLU                  \\ \hline
Hidden layer 3              & 128 neurons, ReLU                 \\ \hline
Optimizer                   & Adam, learning rate = $10^{-3}$   \\ \hline
Loss function               & Cross-entropy                     \\ \hline
Epochs                      & 100                               \\ \hline
\end{tabular}
\end{adjustbox}
\vspace{-10pt}
\end{table}


For comparison purposes, we consider the following benchmark methods:

\begin{enumerate}
\item \textit{SVD perfect channel} based on the singular value
decomposition (SVD) of perfectly known channels under
$3$-bit quantized phase shifter. 


\item \textit{Exhaustive search O-DFT} (with oversampling factor of $\times$4), where the BS exhaustively sweeps
the O-DFT codebook to scan $ N_{\mathrm{BS}} \times$4 narrow beams, and selects
the beam with the highest RSSI value for downlink transmission. 

\item \textit{SHAP FS} where the  SHAP method from XAI is first used for optimizing sensing beam selection by analyzing the contributions of RSSI values across candidate beams. The BS then sweeps a subset of selected wide beams beams determined by the SHAP feature importance rankings to reduce the beam sweeping overhead.

\end{enumerate}

\begin{figure*}[t]
    \centering
    \includegraphics[width=0.8\linewidth]{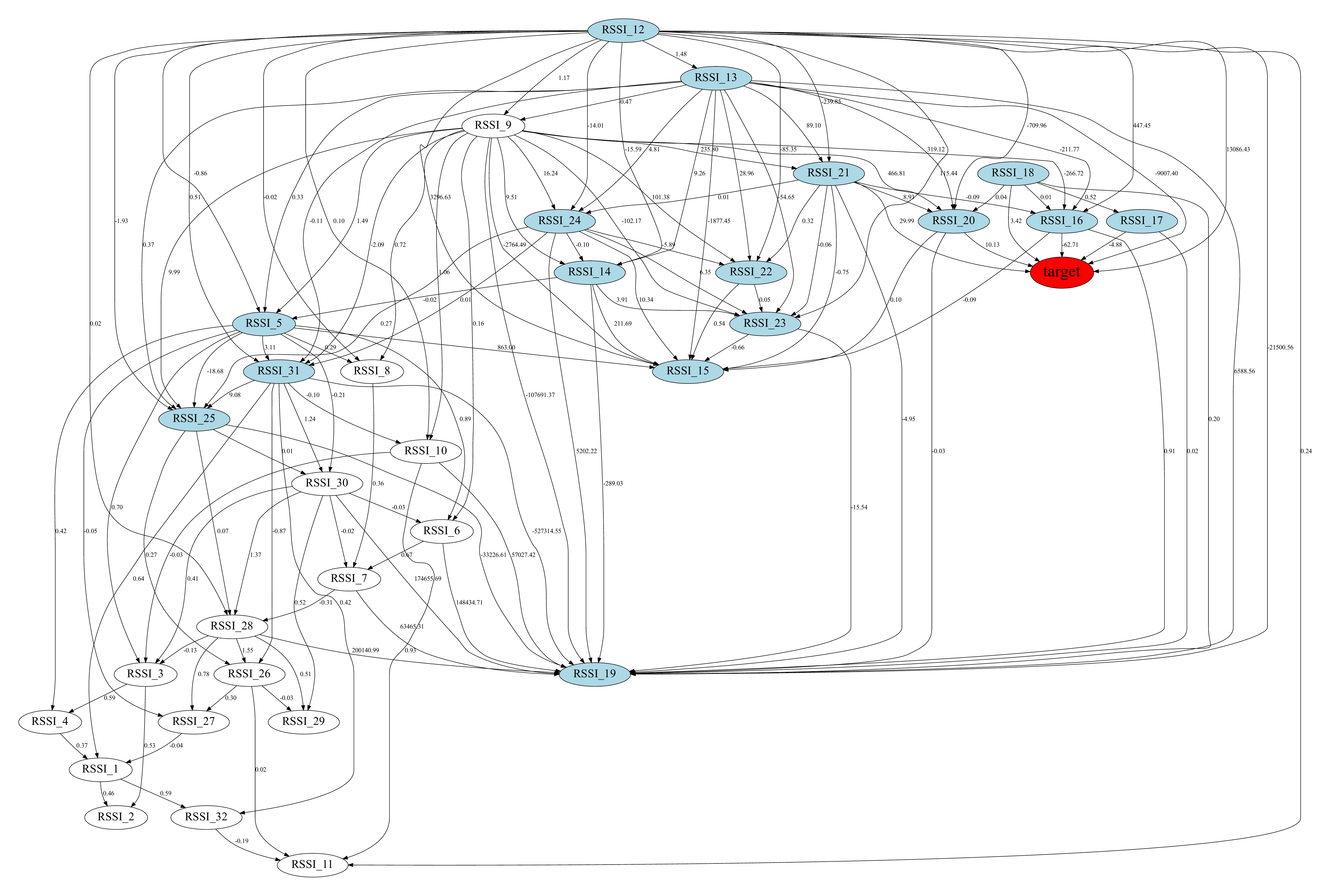}
    \caption{DAG obtained via DirectLiNGAM based causal beam selection approach.}
    \label{fig:CBS}
    \vspace{-8pt}
\end{figure*}

\subsection{Performance metrics}
We consider the following metrics for performance comparison:
\subsubsection{\textit{Top-k} accuracy} This measures the accuracy when the optimal beam index is among the top $k$ predicted beams. {Beam alignment accuracy is improved by sweeping the top $k$ predicted beams, enabling refinement with $k$ additional overhead.}

\subsubsection{Effective Spectral Efficiency} 
This metric quantifies the achievable communication rate while accounting for the overhead introduced by beam alignment. It is defined as the fraction of the frame duration allocated to data transmission, excluding the time spent on initial access (IA) procedures. Mathematically, it is given by:
\begin{equation}
\mathrm{SE} = \frac{T_\mathrm{frame} - T_\mathrm{IA}}{T_\mathrm{frame}} \log_2\left(1 + \mathrm{SNR}\right),
\end{equation}
where $T_\mathrm{frame}$ denotes the total frame duration during which the channel remains coherent. The IA time is expressed as $T_\mathrm{IA} = T_\mathrm{sweep} + T_\mathrm{predict}$, where $T_\mathrm{sweep} = N_b t_s$ is the beam sweep time for scanning $N_b = \widetilde{M}_{\mathrm{w}}$ beams, and $t_s$ is the time to scan a single beam.

\subsection{Performance Evaluation}

\subsubsection{Analysis of Causal Beam Discovery}
Fig. ~\ref{fig:CBS} illustrates the DAG extracted using the DirectLiNGAM algorithm for causal beam selection. Each node in the graph corresponds to either an RSSI input feature or the target beam index, and the directed edges represent the estimated direct causal effects based on residual independence and least-squares regression. As evident in the figure, the most causally influential features with direct edges into the target node include \textit{RSSI\_20}, \textit{RSSI\_16}, \textit{RSSI\_17}, \textit{RSSI\_18}, \textit{RSSI\_19}, and \textit{RSSI\_21}, which are identified as direct parents of the target beam index.  The other features included are based on the total connectivity strength of the variables. Interestingly, the final selection aligns well with the SHAP-based feature importance ranking, supporting the validity of the selected RSSI features for robust and efficient beam prediction.

\begin{figure}[t]
    \centering
    \includegraphics[width=0.8\linewidth]{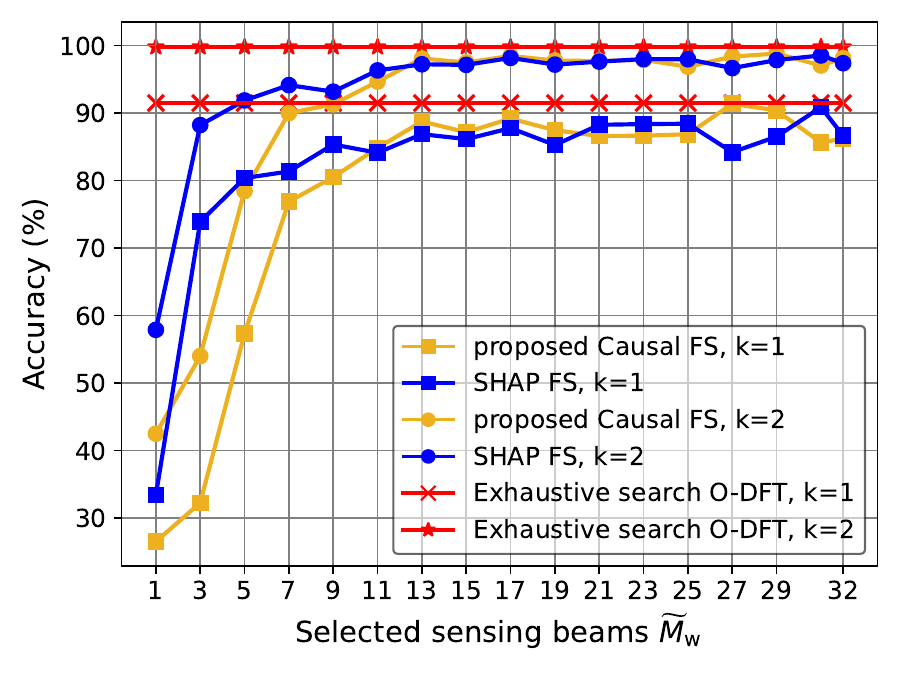}
    \caption{Top-k accuracy performance of different schemes under measurement noise.}
    \label{fig:accuracy_vs_beams}
    \vspace{-15pt}
\end{figure}

\subsubsection{Analysis of Beam Alignment Accuracy} 
Fig.~\ref{fig:accuracy_vs_beams} illustrates the \textit{top-$k = \{1, 2\}$} accuracy as a function of the number of causality-based selected sensing input beam $\widetilde{M}_{\mathrm{w}}$. The exhaustive search approach yields the highest accuracy, primarily due to its reduced vulnerability to noise in beam measurements. Nonetheless, this gain comes at the cost of substantial training overhead and occasional alignment errors arising from noise-induced discrepancies. In contrast, the proposed causal feature selection method achieves 87\% beam alignment accuracy using only $\widetilde{M}_{\mathrm{w}} = 13$ sensing beams, and further improves to 95\% accuracy by exploring an additional $k = 2$ narrow beams. This performance enhancement is largely due to the exclusion of causally irrelevant RSSI features, which would otherwise introduce noise and hinder the learning process.

\subsubsection{Analysis of Effective Spectral Efficiency} Fig.~\ref{fig:SE_vs_features} shows the effective spectral efficiency achieved by the proposed and benchmark schemes across varying numbers of selected features $\widetilde{M}_{\mathrm{w}}$. Unlike the SVD-based and exhaustive search methods, which use fixed input sizes of $32$ and $128$ beams respectively, the proposed approach adaptively selects $\widetilde{M}_{\mathrm{w}}$ based on causal relevance. With just $\widetilde{M}_{\mathrm{w}} = 8$ beams, the proposed \textit{Causal FS} method outperforms the SVD-based solution and matches the performance of the \textit{SHAP FS} approach. Despite its high average SNR, the exhaustive search, requiring all $128$ beams, yields the lowest spectral efficiency. Notably, the spectral efficiency exhibits a unique maximizer $\widetilde{M}^{*}_{\mathrm{w}}$: it increases with $\widetilde{M}_{\mathrm{w}} \leq \widetilde{M}^{*}_{\mathrm{w}}$ due to improved alignment, but declines beyond this point due to overhead and reduced communication time.

\begin{figure}[t]
    \centering
    \includegraphics[width=0.8\linewidth]{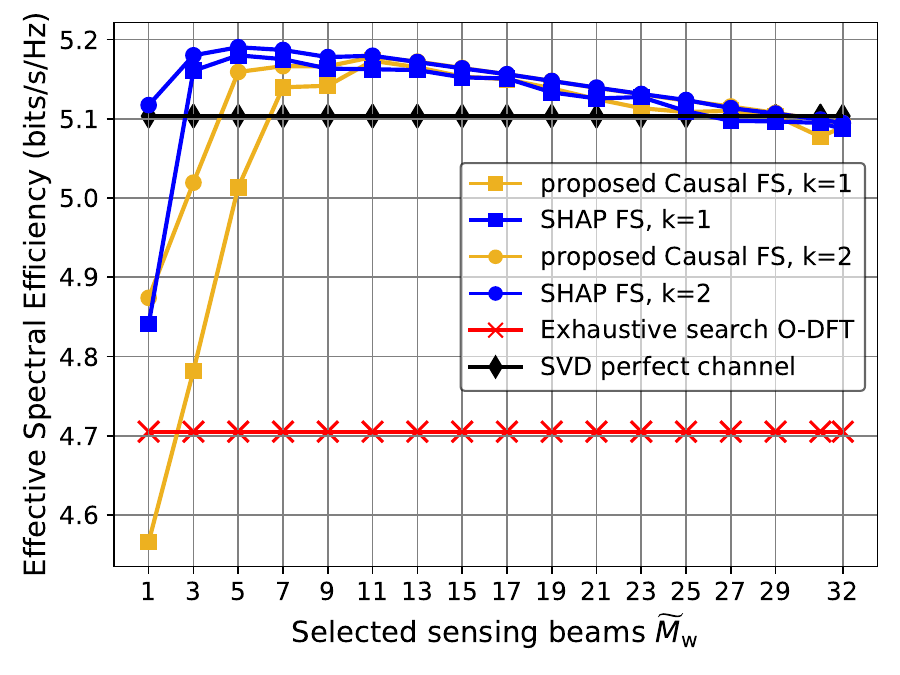}
    \caption{Effective spectral efficiency vs. number of selected sensing beams/features for different approaches.}
    \label{fig:SE_vs_features}
    \vspace{-10pt}
\end{figure}

\subsubsection{Computation, Model, and Time Complexity}
The computational complexity of the proposed DL-based BAE is mainly due to beam sweeping and feedback, both scaling linearly with the number of swept beams. All $N_\mathrm{U}$ UEs simultaneously measure and report RSSIs over $\widetilde{M}_{\mathrm{w}}$ sensing beams during BS sweeping. If the BS additionally sweeps top-$k$ predicted beams per UE, the sweeping complexity becomes $\widetilde{M}_{\mathrm{w}} + N_\mathrm{U} \times k \cdot \mathbbm{1}_{\{k > 1\}}$, and feedback complexity is $N_\mathrm{U} \times \widetilde{M}_{\mathrm{w}} + N_\mathrm{U} \cdot \mathbbm{1}_{\{k > 1\}}$. Model complexity depends on the DNN architecture; with $|\mathbf{x}_S| = \widetilde{M}_{\mathrm{w}}$ input features and hidden layer sizes $H_l$, the parameter count is $\mathcal{O}\left(|\mathbf{x}_S| \cdot H_1 + \sum_{l=1}^3 H_l \cdot H_{l+1}\right)$, and each training update scales as $\mathcal{O}\left(B \cdot \left(|\mathbf{x}_S| \cdot H_1 + \sum_{l=1}^3 H_l \cdot H_{l+1}\right)\right)$ for mini-batch size $B$. Notably, the proposed causal beam selection using DirectLiNGAM significantly reduces runtime to 102.41 seconds seconds compared to 1837.08 seconds for SHAP, without requiring model retraining.

\section{Conclusion}\label{sec:conclusion}
In this paper, we have developed a causally-aware DL framework for efficient beam prediction by introducing a two-stage causal beam selection algorithm. By leveraging causal discovery to learn the dependencies between RSSI inputs and beam decisions, the framework identifies a minimal set of causally relevant features for beam prediction. Simulation results demonstrate that the proposed method achieves 87\% top-1 and 95\% top-2 accuracy with only 13 selected beams, closely matching the exhaustive search performance while significantly reducing beam sweeping overhead. Furthermore, the proposed DirectLiNGAM-based causal beam selection reduces input selection time by 94.4\% compared to SHAP-based methods and decreases beam sweeping overhead by 59.4\% by focusing only on causally relevant features. This approach offers a practical and scalable solution for efficient 6G beam management.

\bibliographystyle{IEEEtran}\bibliography{IEEE_abr,XAI_bibliography}

\begin{thebibliography}{10}
\providecommand{\url}[1]{#1}
\csname url@samestyle\endcsname
\providecommand{\newblock}{\relax}
\providecommand{\bibinfo}[2]{#2}
\providecommand{\BIBentrySTDinterwordspacing}{\spaceskip=0pt\relax}
\providecommand{\BIBentryALTinterwordstretchfactor}{4}
\providecommand{\BIBentryALTinterwordspacing}{\spaceskip=\fontdimen2\font plus
\BIBentryALTinterwordstretchfactor\fontdimen3\font minus \fontdimen4\font\relax}
\providecommand{\BIBforeignlanguage}[2]{{%
\expandafter\ifx\csname l@#1\endcsname\relax
\typeout{** WARNING: IEEEtran.bst: No hyphenation pattern has been}%
\typeout{** loaded for the language `#1'. Using the pattern for}%
\typeout{** the default language instead.}%
\else
\language=\csname l@#1\endcsname
\fi
#2}}
\providecommand{\BIBdecl}{\relax}
\BIBdecl

\bibitem{tutorialBA}
M.~Giordani \emph{et~al.}, ``A tutorial on beam management for {3GPP} {NR} at mm{W}ave frequencies,'' \emph{{IEEE} Commun. Surveys Tuts.}, vol.~21, no.~1, pp. 173--196, 2018.

\bibitem{our_work}
N.~Khan, S.~Coleri, A.~Abdallah, A.~Celik, and A.~M. Eltawil, ``Explainable and robust artificial intelligence for trustworthy resource management in {6G} networks,'' \emph{{IEEE} Commun. Mag.}, vol.~62, no.~4, pp. 50--56, 2024.

\bibitem{XAI1}
A.~Terra, R.~Inam \emph{et~al.}, ``Explainability methods for identifying root-cause of {SLA} violation prediction in 5{G} network,'' in \emph{Proc. IEEE Global Commun. Conf. (GLOBECOM)}.\hskip 1em plus 0.5em minus 0.4em\relax IEEE, 2020, pp. 1--7.

\bibitem{XAI4}
P.~Barnard, N.~Marchetti, and L.~A. DaSilva, ``Robust network intrusion detection through explainable artificial intelligence ({XAI}),'' \emph{IEEE Netw. Lett.}, vol.~4, no.~3, pp. 167--171, 2022.

\bibitem{XAI_ICC}
A.-D. Marcu \emph{et~al.}, ``Explainable artificial intelligence for energy-efficient radio resource management,'' in \emph{Proc. IEEE Wireless Commun. and Netw. Conf. (WCNC)}.\hskip 1em plus 0.5em minus 0.4em\relax IEEE, 2023, pp. 1--6.

\bibitem{lundberg}
S.~M. Lundberg \emph{et~al.}, ``From local explanations to global understanding with explainable {AI} for trees,'' \emph{Nat. Mach. Intell.}, vol.~2, no.~1, pp. 56--67, 2020.

\bibitem{kHAN-TCOM}
N.~Khan, A.~Abdallah, A.~Celik, A.~M. Eltawil, and S.~Coleri, ``Explainable {AI}-aided feature selection and model reduction for {DRL}-based {V2X} resource allocation,'' \emph{{IEEE} Trans. Commun.}, pp. 1--1, 2025.

\bibitem{causal_magazine}
C.~K. Thomas \emph{et~al.}, ``Causal reasoning: Charting a revolutionary course for next-generation ai-native wireless networks,'' \emph{IEEE Veh. Technol. Mag.}, vol.~19, no.~1, pp. 16--31, 2024.

\bibitem{causal_discovery_survey}
A.~Zanga, E.~Ozkirimli, and F.~Stella, ``A survey on causal discovery: Theory and practice,'' \emph{Int. J. Approx. Reason.}, vol. 151, pp. 101--129, 2022.

\bibitem{causal1}
A.~Roy, S.~Banerjee, J.~Sadasivan, A.~Sarkar, and S.~Dey, ``Causally-aware reinforcement learning for joint communication and sensing,'' \emph{IEEE Transactions on Machine Learning in Communications and Networking}, vol.~3, pp. 552--567, 2025.

\bibitem{causal3}
D.~Wheeler and B.~Natarajan, ``Conceptual learning and causal reasoning for semantic communication,'' \emph{IEEE Trans. Cogn. Commun. Netw.}, pp. 1--1, 2025.

\bibitem{causal4}
M.~a. Karaca, ``Utilizing causal learning for cognitive management of {6G} networks,'' in \emph{Proc. IEEE Int. Conf. Mach. Learn. Commun. Netw. (ICMLCN)}, 2024, pp. 234--239.

\bibitem{FastIA}
T.~S.~a. Cousik, ``Deep learning for fast and reliable initial access in {AI}-driven {6G} mm wave networks,'' \emph{IEEE Trans. Netw. Sci. Eng.}, 2022.

\bibitem{KHAN_TWC}
N.~Khan, A.~Abdallah, A.~Celik, A.~M. Eltawil, and S.~Coleri, ``Digital twin-assisted explainable {AI} for robust beam prediction in mm{W}ave {MIMO} systems,'' \emph{{IEEE} Trans. Wireless Commun.}, pp. 1--1, 2025.

\bibitem{directlingam}
S.~Shimizu \emph{et~al.}, ``Directlingam: A direct method for learning a linear non-gaussian structural equation model,'' \emph{J. Mach. Learn. Res.}, vol.~12, no. Apr, pp. 1225--1248, 2011.

\bibitem{alkhateeb2019deepmimo}
A.~Alkhateeb, ``{D}eep{MIMO}: A generic deep learning dataset for millimeter wave and massive {MIMO} applications,'' \emph{arXiv preprint arXiv:1902.06435}, 2019.

\end{thebibliography}

\end{document}